\documentclass[letterpaper, 10 pt, conference]{ieeeconf}  

\IEEEoverridecommandlockouts                              

\usepackage{graphicx}
\usepackage{amssymb}
\usepackage{booktabs}
\usepackage{amsmath}
\usepackage{paracol}
\usepackage{empheq}
\usepackage[table]{xcolor}
\usepackage{tabularx}
\usepackage{multirow}
\usepackage[pagebackref,breaklinks,colorlinks]{hyperref}
\usepackage{paracol}
\usepackage[capitalize]{cleveref}
\crefname{section}{Sec.}{Secs.}
\Crefname{section}{Section}{Sections}
\Crefname{table}{Table}{Tables}
\crefname{table}{Tab.}{Tabs.}
\usepackage{tikz,xcolor}
\usepackage{ragged2e}
\usepackage{balance}

\definecolor{lime}{HTML}{A6CE39}
\DeclareRobustCommand{\orcidicon}{%
    \begin{tikzpicture}
    \draw[lime, fill=lime] (0,0) 
    circle [radius=0.16] 
    node[white] {{\fontfamily{qag}\selectfont \tiny ID}};
    \draw[white, fill=white] (-0.0625,0.095) 
    circle [radius=0.007];
    \end{tikzpicture}
    \hspace{-2mm}
}

\newcommand{\ibrahim}{\href{https://orcid.org/0000-0002-5376-2477}{\orcidicon}}
\newcommand{\naveed}{\href{https://orcid.org/0000-0003-3406-673X}{\orcidicon}}
\newcommand{\haitian}{\href{https://orcid.org/0009-0000-7544-4667}{\orcidicon}}
\newcommand{\ajmal}{\href{https://orcid.org/0000-0002-5206-3842}{\orcidicon}}

\title{\LARGE \bf
Multistream Network for LiDAR and Camera-based 3D Object Detection in Outdoor Scenes
}

\author{Muhammad Ibrahim\ibrahim$^{1}$, Naveed Akhtar\naveed$^{2}$, Haitian Wang\haitian$^{1}$,  Saeed Anwar$^{1}$, and  Ajmal Mian\ajmal$^{1}$ 
\thanks{Ajmal Mian is the recipient of an Australian Research Council Future Fellowship Award (project \# FT210100268)
and Naveed Akhtar is a recipient of the Australian Research Council Discovery Early Career Researcher Award (project number DE230101058), both funded by the Australian Government.}
\thanks{$^{1}$Department of Computer Science, The University of Western Australia.
        {\tt (muhammad.ibrahim@, 23815631@student., saeed.anwar@, ajmal.mian@) uwa.edu.au}
        }%
 \thanks{$^{2}$ School of Computing \& Information Systems, The University of Melbourne, 
         {\tt\small naveed.akhtar1@unimelb.edu.au}}}

\begin{document}

  \makeatletter
\let\@oldmaketitle\@maketitle
\renewcommand{\@maketitle}{\@oldmaketitle  

    }

\maketitle 

\thispagestyle{empty}
\pagestyle{empty}

\begin{abstract}
Fusion of LiDAR and RGB data has the potential to enhance outdoor 3D object detection accuracy. To address real-world challenges in outdoor 3D object detection, fusion of LiDAR and RGB input has started gaining traction. However, effective integration of these modalities for precise object detection task still remains a largely open problem. To address that, we propose a MultiStream Detection (MuStD) network, that meticulously extracts task-relevant information from both data modalities. The network follows a three-stream structure. Its LiDAR-PillarNet stream extracts sparse 2D pillar features from the LiDAR input while the LiDAR-Height Compression stream computes Bird's-Eye View features. An additional 3D Multimodal  stream  combines RGB and LiDAR features using UV mapping and polar coordinate indexing. Eventually, the features containing comprehensive spatial, textural and geometric information are carefully fused and fed to  a detection head for 3D object  detection. Our extensive evaluation on the challenging KITTI Object Detection Benchmark using \href{https://www.cvlibs.net/datasets/kitti/eval_object_detail.php?&result=d162ec699d6992040e34314d19ab7f5c217075e0}{public testing server} establishes the efficacy of our method by achieving new state-of-the-art or highly competitive results in different categories while remaining among the most efficient methods.  Our code will be released through \href{https://github.com/IbrahimUWA/MuStD.git}{MuStD GitHub repository}.

\end{abstract}

\section{INTRODUCTION}
Accurate outdoor 3D object detection is crucial for reliable autonomous navigation~\cite{Chen2017MV3D}, \cite{arnold2019survey}. Currently, LiDAR stand out as the primary sensor to enable that~\cite{prakash2021multi}, \cite{wu2022sparse}, \cite{wu2022sfd}. However, sampling sparsity and  partial measurements caused by occlusion compromise 3D objection detection using only the LiDAR data~\cite{wu2023virtual}, \cite{9826439}, \cite{yang2022graph}.
This limitation can potentially be addressed by fusing LiDAR measurements with RGB camera inputs~\cite{zhu2023vpfnet}, \cite{bai2022transfusion}, \cite{guan2021m3detr}. The latter are known for accurately acquiring high-resolution texture information, which complements LiDAR input for improved 3D object detection.

Currently, fusion of LiDAR and RGB data for 3D object detection is gaining rapid traction~\cite{zhu2023vpfnet}, \cite{bai2022transfusion}, \cite{meng2024efficient}, \cite{guan2021m3detr}. 
Whereas early methods have typically employed late fusion at regions of interest or Bird’s Eye View (BEV) \cite{bai2022transfusion}, \cite{meng2024efficient}, the more recent approaches use depth completion to enrich the LiDAR’s sparse data with pseudo points~\cite{tian2023acfnet}, \cite{guan2021m3detr}. The virtual points help mitigating the sparsity issue, particularly for distant and occluded objects, resulting in improved geometric details for object detection.

\begin{figure*}
\begin{center}
\includegraphics[width=1\textwidth]{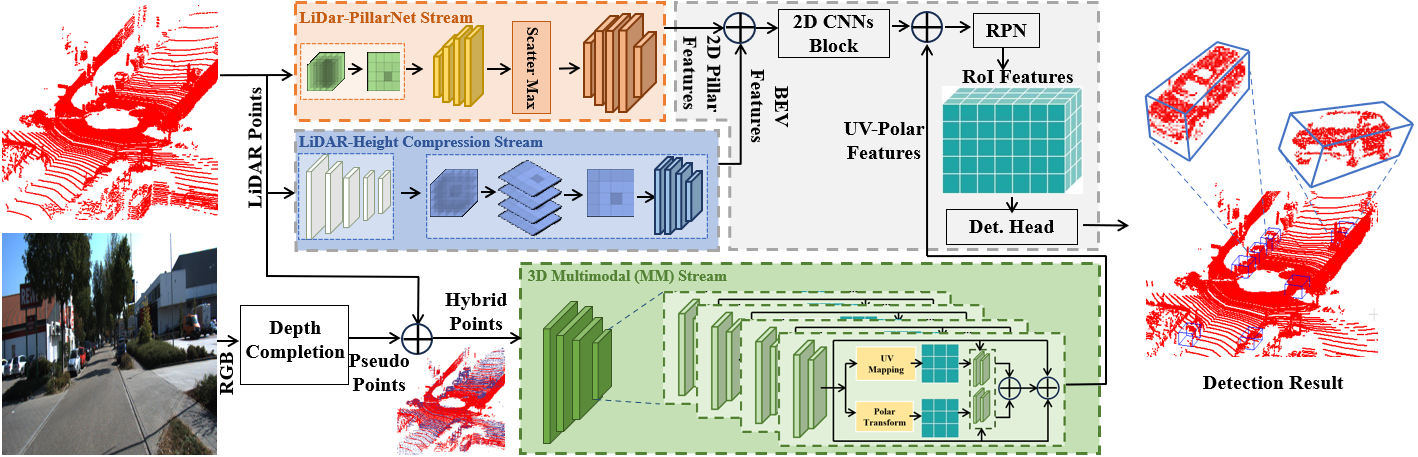}
\end{center}
\vspace{-3mm}
\caption{Overview of the proposed MuStD network for 3D object detection comprising three data processing streams. (1) The PillarNet Stream converts LiDAR frames into 2D Pillar features. It leverages an MLP, Scatter Max layer and a 2D sparse convolution for  feature extraction. (2) The LiDAR-Height Compression Stream uses 3D sparse convolutions on LiDAR input to extract 3D features, which are then transformed to bird's eye view (BEV).
(3) The 3D Multimodal (MM) Stream processes both RGB and LiDAR data, where the former is mapped  to  pseudo points. The featuring MM layers exploit UV mapping and Polar transforms to extract 2D and 3D sparse features. These UV-Polar features are combined with the features resulting from the fusion of the other stream features and processed by a Detection Head for object detection.}
\label{fig:main}
\vspace{-5mm}
\end{figure*}

Commonly, contemporary fusion-based 3D detection methods rely on isolated strategies, such as UV mapping or polar transformation~\cite{wu2023virtual, zhu2023vpfnet, bai2022transfusion}. The former is effective in aligning LiDAR points with 2D images; however, it fails to adequately encode the spatial relationships of point clouds, leading to the underutilization of critical geometric details in the scene. Similarly, while polar transformation is proficient at encoding spatial orientation and distance information, it is ineffective in integrating the rich semantic and texture information provided by RGB images~\cite{wang2021multi, caesar2020nuscenes, 8943956}. These isolated approaches fail to effectively merge LiDAR’s 3D spatial context with RGB’s dense semantic information, resulting in suboptimal detection performance, especially in complex scenarios involving occluded or distant objects.

In this work, we propose a MultiStream Detection (MuStD) network that effectively integrates LiDAR point clouds with RGB images for enhanced 3D object detection. Illustrated in Fig.~\ref{fig:main}, our technique addresses  multimodal fusion by optimizing geometric and spatial information extraction and improving object orientation estimation. Our approach is structured as three parallel data processing streams, where each stream eventually contributes to the fusion of rich geometric details of LiDAR input and textural information of RGB images. Our contributions are as follows.
\vspace{-0.3mm}
\begin{itemize}
    \item \textbf{Multistream Detection (MuStD) Network}: We propose the MuStD network containing three parallel streams combining LiDAR and RGB data for enhanced 3D object detection. MuStD  effectively leverages the strengths of each modality - LiDAR for spatial geometry and RGB for texture detail - while mitigating their limitations through comprehensive feature fusion. 
    
    \item \textbf{3D Multimodal (MM) stream}: A central innovation in our MuStD network is the 3D MM stream, which effectively integrates UV mapping and polar coordinate indexing. UV mapping aligns 3D LiDAR points with 2D image features to capture fine-grained texture and appearance details from the RGB modality. Concurrently, polar coordinate indexing encodes spatial orientation and depth relationships in the scene, enhancing the geometric representation of objects. 
      \item \textbf{UV-Polar block}: We propose a novel block that projects 3D sparse convolution features onto a UV image and polar space, creating 2D grid representations. These are processed with 2D sparse convolutions and then merged with the original 3D sparse features, resulting in a unified feature set that integrates both position and orientation information of objects.

\end{itemize}
\vspace{-0.3mm}

Extensive experiments show that the proposed MuStD network achieves state-of-the-art or highly competitive results on the KITTI object detection leaderboard while remaining among the most efficient methods. 

\vspace{-1mm}
\section{Related Work}
\vspace{-1mm}

In 3D detection, LiDAR-only methods still remain popular due to the spatial precision of LiDAR input~\cite{ICRAencode}, \cite{ICRAmatching}, \cite{ICRAlidar}, \cite{ICRAframe}, \cite{CVPR2021_Offboard3D}, 
Techniques such as PointPillars \cite{Lang2019PointPillars} and SECOND \cite{Yan2018SECOND}, perform real-time processing by converting point clouds into 2D pseudo-images and then apply 2D convolutions. This reduces computational cost while maintaining detection accuracy. Advanced models like PV-RCNN \cite{shi2020pv} and Voxel-RCNN \cite{deng2021voxel} enhance performance by combining point-wise and voxel-wise features using sparse 3D convolutions and attention mechanism, achieving improved results.

Recent research has also focused on optimizing the integration of LiDAR and RGB data to improve 3D detection~\cite{ji2022vision}. Fusing the two modalities aims to exploit precise geometric information from the LiDAR input and dense textural details from the images. Early fusion methods like AVOD~\cite{Ku2018AVOD} and MV3D~\cite{Chen2017MV3D} fuse LiDAR and image features after their independent extraction. More recently, techniques such as 3D-CVF~\cite{yoo20203d} align features from both modalities at the feature level. To address the sparsity of LiDAR input, methods such as SFD~\cite{wu2022sfd} and VirConv~\cite{VirConv} use point cloud completion to generate dense pseudo points. However, this also introduces noise, particularly at object boundaries. VPFNet~\cite{9826439} refines feature fusion by selectively combining features based on spatial reliability, mitigating noise and improving detection performance.

Multimodal fusion for 3D object detection mainly faces challenges in aligning features across modalities and suppressing noise~\cite{Sensors2024, sensors2022}. NRConv~\cite{wang2023nrconv} addresses this with noise-resistant convolutions that enhance feature extraction by reducing the influence of noisy data. The 2DPASS framework~\cite{ye2021twopass} improves segmentation and detection by integrating 2D semantic priors into 3D point clouds. Other advances focus on handling the complexities of real-world scenarios. For instance, \cite{zhou2023ted} employs transformation-equivariant convolutions to improve robustness against rotation and reflection variations in autonomous driving. Similarly, Graph-VoI~\cite{yang2022graph} uses graph neural networks to model complex object relationships, leveraging both geometric and semantic features. Despite advances in combining RGB and LiDAR modalities, effective outdoor 3D object detection remains an open problem.
We integrate advanced mapping and efficient convolutions while optimizing multimodal fusion to achieve state-of-the-art performance for this task. 


\section{Methodology}
\vspace{-1mm}
Figure~\ref{fig:main} illustrates the schematics of our MultiStream Detection (MuStD) network. The proposed network utilizes three parallel data processing streams. 
(a) 3D Multimodal (MM) stream, which 
processes 3D hybrid points using our proposed UV-polar block at each layer.
(b) The LiDAR-Height Compression stream that extracts sparse 3D features from LiDAR frames, compresses them using a height module, and processes them with 2D CNN blocks to capture spatial relationships and object geometry. 
 (c) The LiDAR-PillarNet stream, which converts 3D point clouds into 2D polar representations through pillar-based voxelization and neural networks, effectively extracting robust geometric features such as object orientation and localization.
 Details of the MuStD network are given below.

\vspace{-0.7mm}
\subsection{3D Multimodal (MM) stream}
\vspace{-0.7mm}

\begin{figure}[t]
    \centering
    \includegraphics[width=1\linewidth]{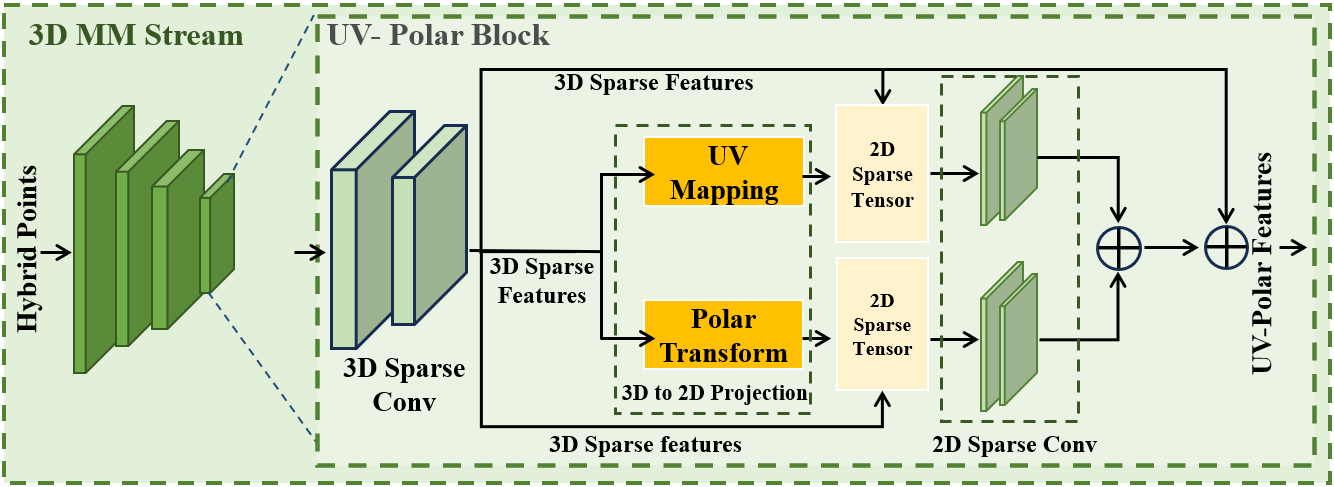}
    \vspace{-7mm}
    \caption{3D MM stream of MuStD network integrates UV mapping and polar coordinate indexing. UV mapping aligns 3D LiDAR points with 2D image features, capturing texture and appearance, while polar coordinate indexing encodes orientation and depth for improved geometric representation.
    }
    \label{fig:streamC}
    \vspace{-4mm}
\end{figure}

The 3D MM stream (see Fig.~\ref{fig:streamC}), a key innovation of the MuStD Network, is specifically designed to enhance 3D object detection by integrating LiDAR and camera data. The novelty lies in its ability to extract 3D sparse features that include both UV and polar information, crucial for accurately determining an object's location and orientation. This stream is composed of a series of UV-Polar blocks with channel sizes of 16, 32, 64, and strides of 1, 2, 2, 2, processing hybrid 3D points through the stages discussed below.

\noindent \textbf{Hybrid Points Generation}: 
In this step, pseudo point clouds are first generated from RGB images using point cloud completion~\cite{wu2022sparse}. To improve efficiency, about 80\% of these pseudo points are discarded and the remaining points are fused with LiDAR point clouds to create Hybrid 3D points with enriched spatial and geometric information. The hybrid points are then processed by the 3D MM stream.

\noindent \textbf{UV-Polar Block}: This block serves as the foundational unit of the MM stream, designed to capture detailed geometric, orientation, and spatial features of objects. Initially, it processes the hybrid points using 3D sparse convolutional layers to extract high-level 3D sparse features. These features are then projected onto image and polar spaces through two parallel transformations: UV Mapping and Polar Transform.

The UV Mapping process projects 3D points to a 2D plane, aligning them with their corresponding RGB image features. This is achieved by calculating the UV coordinates \( (u, v) \) as \( \left( \frac{x}{z}, \frac{y}{z} \right) \), where \( x \), \( y \), and \( z \) 
represent the 3D coordinates of each point. The input 3D sparse feature, originally of size \(F_{3D} \in \mathbb{R}^{H \times W \times D \times C} \), is projected onto a 2D feature space of size \( F_{UV} \in \mathbb{R}^{1600 \times 600 \times C} \). The projected 2D features are then processed through a series of 2D sparse convolutional layers to capture texture-rich details, enabling the network to identify fine-grained 2D features crucial for accurate object detection.

The Polar Transform projects the 3D sparse features to polar coordinates \( F_{P} \in \mathbb{R}^{1600 \times 600 \times C} \). These polar 2D features are then processed through 2D sparse convolutional layers to extract critical information related to the orientation and distance of objects, enhancing the network's ability to detect and localize objects within the scene. The conversion is defined as
\vspace{-1mm}
\begin{equation*}
(r, \theta, \phi) = (\sqrt{x^2 + y^2 + z^2}, \tan^{-1} (\frac{y}{x} ), \tan^{-1}(\frac{z}{\sqrt{x^2 + y^2}}) ),
\end{equation*}
where $r$ is the radial distance, $\theta$ is the azimuth and $\phi$ is the elevation angle. This representation helps manage variations in scale and rotation commonly found in real-world data. 

Finally, polar, UV, and input 3D sparse features are fused within the block to form a comprehensive feature set, $F_{MM}$ at each layer of the stream. The equation below  
represents $F_{MM}$ for the $l+1^{\text {th}}$ layer, where $\mathbf{X}$, $\mathbf{W}$, $\mathcal{U}$, $\mathcal{P}$, $\oplus$, $\circledast$, and $\star$ denote input feature, kernel weights, UV mapping, polar transformation, concatenation, 2D sparse convolution and 3D sparse convolution operations respectively.

\vspace{-4mm}
\begin{equation*} 
\text{\small $
\mathbf{F}_{\text{MM}}^{(l+1)} = \left( \mathbf{X} \star \mathbf{W} \right)
\oplus \left( \mathcal{U} \left( \mathbf{X} \right) \circledast \mathbf{W} \right) 
\oplus \left( \mathcal{P} \left( \mathbf{X} \right) \circledast \mathbf{W} \right).
$}
\end{equation*}

\begin{figure}[t]
    \centering
    \includegraphics[width=1\linewidth]{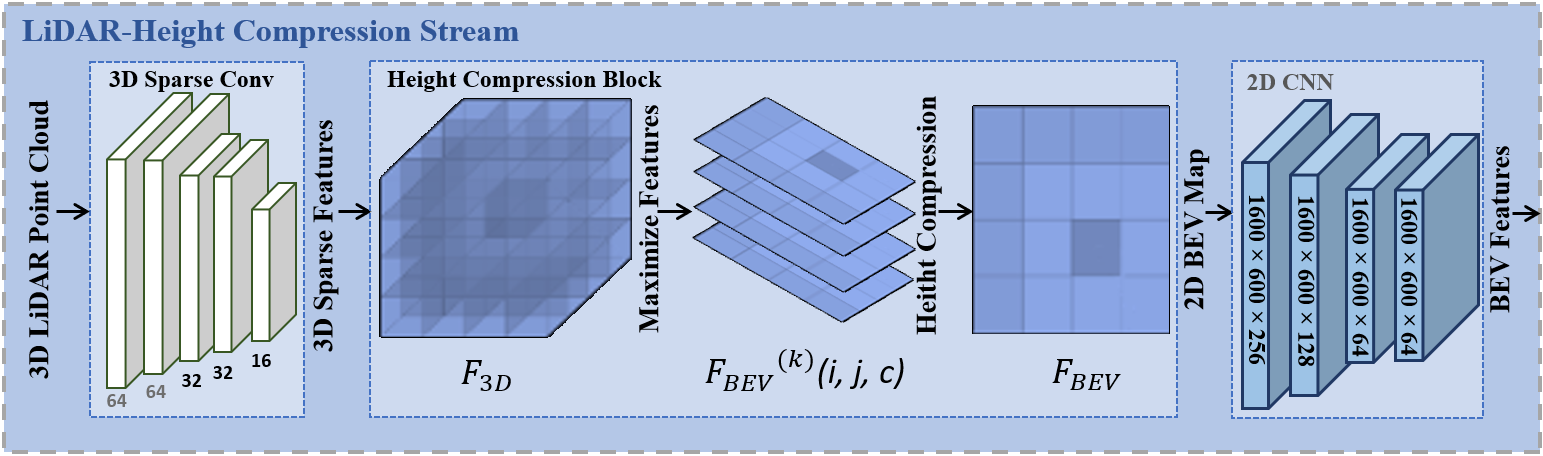}
    \vspace{-6mm}
    \caption{LiDAR-Height Compression stream processes LiDAR points with 3D sparse CNN to extract 3D geometric feature maps which are then aggregated by a height compression block along the z-axis to get 2D feature maps for processing by a 2D CNN to reduce computational complexity. The figure is best viewed enlarged for architectural details.}
    \label{fig:streamB}
    \vspace{-3mm}
\end{figure}

\vspace{-4mm}
\subsection{LiDAR Height Compression stream}
The LiDAR Height Compression stream is designed to efficiently extract geometric features from the raw LiDAR point cloud. It consists of 3D sparse convolution blocks, height compression followed by a series of 2D CNN layers. 

\noindent \textbf{3D Sparse Convolution Blocks}: The raw LiDAR data is processed through a sequence of 3D Sparse Convolution blocks to effectively handle its high dimensionality and sparsity. 
The LiDAR frame is first voxelized and then passed through 3D sparse convolutional blocks with channel sizes of 16, 32, 32, 64, 64, and strides 1, 2, 2, 2, 1, respectively. Each block contains three convolution layers, with the first layer downsampling to capture essential geometric details. The final output is a 3D feature tensor \( F_{3D} \in \mathbb{R}^{H \times W \times D \times C} \), where \( H \), \( W \), \( D \), and \(C\) represent the height, width, depth, and the number of feature channels, respectively.

\noindent \textbf{Height Compression Block}: To reduce computational complexity while retaining vital spatial information, the next stage involves compressing the height dimension of the 3D feature tensor \( F_{3D} \) (see Fig.~\ref{fig:streamB}). This block focuses on capturing the spatial relationships and geometry of objects by processing height information. Height compression aggregates features along the z-axis, effectively projecting the 3D feature map onto a 2D plane, forming the BEV feature map \( F_{BEV} \) \cite{jhaldiyal2023semantic}. This process is defined as
\vspace{-2mm}
\begin{equation*}
\text{\small $
F_{BEV}(i, j, c) = \max_{k \in [1, H]} F_{3D}(i, j, k, c),
$}
\end{equation*}
where \( F_{BEV} \in \mathbb{R}^{W \times D \times C} \) is the compressed feature map that retains the most prominent features (maximum values) along the height dimension. The BEV feature map \( F_{BEV} \) is then refined through a series of 2D CNN layers, producing the processed features \( \mathbf{F}_{\text{BEV,2D}} \) as 

\vspace{-5mm}
\begin{equation*}
\text{\scriptsize $
\mathbf{F}_{\text{BEV,2D}}^{(l+1)} = \sigma \left( \sum_{k=1}^{K} \mathbf{W}^{(l,k)} \ast \mathbf{F}_{\text{BEV}}^{(l)} + \mathbf{b}^{(l)} \right), \quad l = 1, 2, \dots, L.
$}
\end{equation*}
Here, \( \mathbf{F}_{\text{BEV,2D}}^{(l+1)} \in \mathbb{R}^{H' \times W' \times C'} \) is the refined 2D BEV feature map after the \( l \)-th convolutional layer. \( \mathbf{W}^{(l,k)} \in \mathbb{R}^{K \times K \times C \times C'} \) is the learnable kernel for the \( l \)-th layer, where \( K \) represents the kernel size and \( k \) indexes the filters. \( \mathbf{b}^{(l)} \in \mathbb{R}^{C'} \) is the bias term. \( \sigma(\cdot) \) represents the non-linear activation function ReLU. \( \ast \) denotes the 2D convolution operation.

\subsection{LiDAR-PillarNet Stream}

\begin{figure}
    \centering
    \includegraphics[width=1\linewidth]{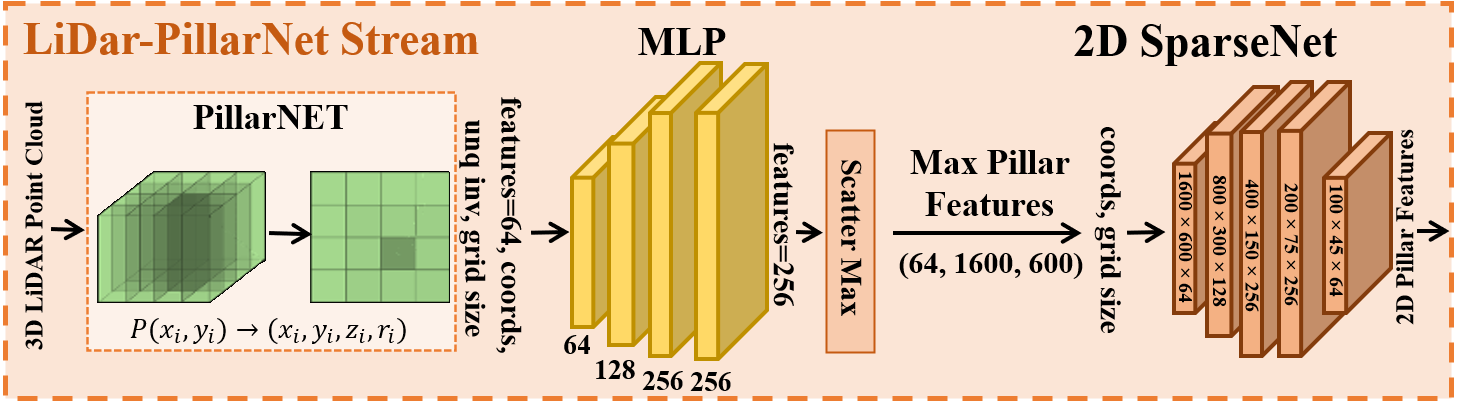}
    \vspace{-7mm}
    \caption{The LiDAR-PillarNet architecture voxelizes raw LiDAR data into 2D pillar features, refines them with a Multi-Layer Perceptron (MLP) to extract Max Pillar Features, and processes these features through a 2D SparseNet for further refinement. Best viewed enlarged.}
    \label{fig:streamA}
    \vspace{-5mm}
\end{figure}

The LiDAR-PillarNet Stream (see Fig.~\ref{fig:streamA}) converts 3D point clouds into 2D representations using pillar-based voxelization followed by feature extraction that extracts robust geometric features for high detection accuracy, while reducing computational complexity. 
The pillar-based voxelized data is passed through a Multi-Layer Perceptron (MLP) followed by 2D sparse convolutions to extract features. 
This stream initially converts raw LiDAR point cloud data into vertical columns or ``pillars''. Each pillar aggregates LiDAR points based on their \(x\) and \(y\) coordinates, effectively discretizing the continuous 3D space into a grid representation.

\begin{equation*}\label{eq:pillar}
\text{\small $
P(x, y) = \left\{ (x_i, y_i, z_i, r_i) \mid \left\lfloor \frac{x_i}{\Delta x} \right\rfloor = x, \left\lfloor \frac{y_i}{\Delta y} \right\rfloor = y \right\}
$},
\end{equation*}
where, \((x_i, y_i, z_i, r_i)\) represent the coordinates and reflectance of a LiDAR point, and \(\Delta x\) and \(\Delta y\) are the pillar sizes in the \(x\) and \(y\) dimensions. This step generates a sparse pseudo-image, where each cell in the 2D grid corresponds to a pillar containing point cloud information~\cite{li2023tinypillarnet}. After discretization, the stream encodes each point within its pillar into a fixed-size feature vector relative to the pillar's center, effectively capturing both local and global spatial context.

The encoded features are refined through an MLP to capture the most relevant aspects for object detection. A Scatter Max layer then selects key features and aggregates them into a 2D grid, preserving the spatial arrangement of the pillars. This grid is processed by a 2D SparseNet, designed to handle data sparsity while extracting high-level geometric features, producing the output ${\mathbf{F}}_{\mathcal{S},\text{2D}}$.

\vspace{-4mm}
\begin{equation*}
\text{\small $
\mathbf{F}_{\mathcal{S}, \text{2D}}^{(l+1)} = \sigma \left( \sum_{k=1}^{K} \mathbf{W}^{(l,k)} \circledast  \mathbf{F}_{\mathbf{max}}^{(l)} + \mathbf{b}^{(l)} \right), \quad l = 1,2, \dots, L.
$}
\end{equation*}

where \( \mathbf{F}_{\mathcal{S}, \text{2D}}^{(l+1)} \in \mathbb{R}^{H \times W \times C'} \) is the output feature of PillarNet map after the \( l \)-th 2D sparse convolutional layer, \( \mathbf{W}^{(l,k)} \in \mathbb{R}^{K \times K \times C \times C'} \) is the learnable kernel matrix for the \( k \)-th filter in the \( l \)-th layer, where \( K \) is the kernel size, and \( \mathbf{b}^{(l)} \in \mathbb{R}^{C'} \) is the bias term for the \( l \)-th layer. The function \( \sigma(\cdot) \) is the ReLU non-linear activation function and \( \circledast \) denotes the sparse convolution operation. This formulation refines the feature map \( \mathbf{F}_{\mathcal{S}, \text{2D}}^{(l)} \), capturing the essential spatial and geometric properties of the scene. 

\makeatletter
\def\thickhline{%
  \noalign{\ifnum0=`}\fi\hrule \@height \thickarrayrulewidth \futurelet
  \reserved@a\@xthickhline}
\def\@xthickhline{\ifx\reserved@a\thickhline
              \vskip\doublerulesep
              \vskip-\thickarrayrulewidth
             \fi
      \ifnum0=`{\fi}}
\makeatother
\newlength{\thickarrayrulewidth}
\setlength{\thickarrayrulewidth}{2\arrayrulewidth}

\vspace{-1mm}

\subsection{Feature Fusion and Object Detection}
At the final stage, features from the LiDAR-PillarNet, LiDAR-Height Compression, and 3D MM streams are integrated for robust object detection. The features from the LiDAR streams are concatenated and processed through 2D CNN layers, creating a unified feature set. These are then combined with the 3D MM stream features to generate enhanced feature maps. The final feature map, denoted as \( F_{\text{H}} \in \mathbb{R}^{H' \times W' \times C'} \), is derived by fusing the 2D and 3D feature maps. The fusion is performed as follows.
\begin{equation*}
\text{\small $
\mathbf{F}_{\text{H}} = \sigma \left( \sum_{l=1}^{L} \mathbf{W}^{(l,k)} \circledast \left( \mathbf{F}_{\text{BEV,2D}} \oplus 
\mathbf{F}_{\mathcal{S}, \text{2D}} \right) \right) \oplus \mathbf{F}_{\text{MM}}
$}.
\end{equation*}

The final feature map is then processed by the Region Proposal Network (RPN) to generate candidate Regions of Interest (RoI) as bounding boxes. These proposals are then converted into fixed-size feature maps via RoI pooling. The detection head, consisting of fully connected layers, processes these RoI features to output class scores and refined bounding box coordinates as \((C_{\text{obj}}, B_{\text{refined}}) = \text{Det\_Head}(F_{\text{RoI}}) \). Finally, non-maximum suppression (NMS) is applied to remove redundant boxes for precise object detection.

\begin{table}[t]
    \centering
    \vspace{-5mm}
\caption{Server-generated object detection results (average Precision in \%) for 2D, 3D, BEV, and car orientation on the KITTI test set for Easy, Moderate, and Hard categories.}
\vspace{-1mm}
\begin{tabular}{l | c | c  |c | c}
\thickhline
{\bf Benchmark} & {\bf Easy} & {\bf Moderate} & {\bf Hard} & {\bf mean AP} \\ \hline
Car (Detection) & 97.91 & 97.21 & 94.04 & 96.39  \\
Car (Orientation) & 97.88 & 97.03  & 93.74  & 96.22  \\
Car (3D Detection) & 91.03  & 84.36  & 80.78 & 85.39  \\
Car (Bird's Eye View) & 94.62  & 91.13  & 88.28  & 91.34 \\
\thickhline
\end{tabular}

\label{tab:table6}
\vspace{-2mm}
\end{table}

\begin{table*}[t!]
\caption{Results on the KITTI test set generated by the online server. Average precision (AP) and average orientation similarity (AOS) in \% are reported for 2D car detection and orientation, respectively. Best results are bolded and 2nd best underlined.} 
\vspace{-3mm}
\centering
\begin{tabular}{l|c|c c c c|c c c c| c}
\thickhline
 \multirow{2}{4em}{Approach} &  \multirow{2}{4em}{Reference}    &  \multicolumn{4}{c|}{Car 2D AP} &   \multicolumn{4}{c|}{Car Orientation AOS} &  \multirow{2}{5em}{Time (ms)}  \\   \cline{3-10}  
&   & Easy & Moderate & Hard & mean AP &  Easy & Moderate & Hard & mean AP &  \\ \hline  \hline 

MVRA-FRCNN~\cite{Choi_2019_ICCV} & ICCV 2019 & 95.87 & 94.98 & 82.52 & 91.12 & 95.66 & 94.46 & 81.74 & 90.62 & 180 \\
CLOCs~\cite{pang2020CLOCs} & IROS 2020  & 96.77 & 96.07 & 91.11 & 94.65 & 96.77 & 95.93 & 90.93 & 94.54 & 100 \\

SPA-Net~\cite{ye2021spanet} & PRICAI 2021 & 96.54 & 95.46 & 90.47 & 94.16 & 96.31 & 95.03 & 89.99 & 93.78 & 60 \\
VoTr-TSD~\cite{mao2021votr} & ICCV 2021 & 95.97 & 94.94 & 92.44 & 94.45 & 95.95 & 94.81 & 92.24 & 94.33 & 70 \\
Pyramid R-CNN~\cite{mao2021pyramid} & ICCV 2021 & 95.88 & 95.13 & 92.62 & 94.54 & 95.87 & 95.03 & 92.46 & 94.45 & 70 \\

3D D-Fusion~\cite{kim20223d} & ArXiv 2022 & 96.54 & 95.82 & 93.11 & 95.16 & 96.53 & 95.76 & 93.01 & 95.10 & 100 \\
CasA~\cite{casa2022} & TGRS 2022 & 96.52 & 95.62 & 92.86 & 94.99 & 96.51 & 95.53 & 92.71 & 94.92 & 100 \\
VPFNet~\cite{9826439} & TMM 2022 & 96.64 & 96.15 & 91.14 & 94.64 & 96.63 & 96.04 & 90.99 & 94.55 & \underline{60} \\
Graph-VoI~\cite{yang2022graphrcnn} & ECCV 2022 & 96.81 & \underline{96.38} & 91.20 & 94.80 & 96.81 & \underline{96.29} & 91.06 & 94.72 & 70 \\
SFD~\cite{wu2022sparse} & CVPR 2022 & \textbf{98.97} & 96.17 & 91.13 & 95.42 & \textbf{98.95} & 96.05 & 90.96 & 95.32 & 100 \\

DVF (Voxel-RNN)~\cite{mahmoud2022dense} & WACV 2023 & 96.60 & 95.77 & 90.89 & 94.42 & 96.59 & 95.63 & 90.71 & 94.31 & 100 \\
OcTr~\cite{zhou2023octr} & CVPR 2023 & 96.48 & 95.84 & 90.99 & 94.44 & 96.44 & 95.69 & 90.78 & 94.30 & \underline{60} \\
Focals Conv~\cite{focalsconv-chen} & CVPR 2023 & 96.30 & 95.28 & 92.69 & 94.76 & 96.29 & 95.23 & 92.60 & 94.71 & 100 \\
TED~\cite{TED} & AAAI 2023 & 96.64 & 96.03 & {93.35} & {95.34} & 96.63 & 95.96 & \underline{93.24} & {95.28} & 100 \\
MLF-DET~\cite{lin2023mlf} & ICANN 2023 & 96.89 & 96.17 & 88.90 & 93.99 & 96.87 & 96.09 & 88.78 & 93.91 & 90 \\
PVT-SSD~\cite{yang2023pvt-ssd} & CVPR 2023 & 96.75 & 95.90 & 90.69 & 94.45 & 96.74 & 95.83 & 90.58 & 94.38 & \textbf{50} \\
VirConv-T~\cite{VirConv} & CVPR 2023 & \underline{98.93} & \underline{96.38} & \underline{93.56} & \underline{96.29} & \underline {98.64} & 96.01 & {93.12} & \underline{95.92} & 90 \\ \hline \hline
Ours & - & {97.91} & \textbf{97.21} & \textbf{94.04} & \textbf{96.39} & 97.88 & \textbf{97.03} & \textbf{93.74} & \textbf{96.22} & \textbf{50} \\

\thickhline
\end{tabular}
\label{tab:table5}
\vspace{-6mm}
\end{table*}

\begin{figure}
\centering
\includegraphics[width=1\linewidth]{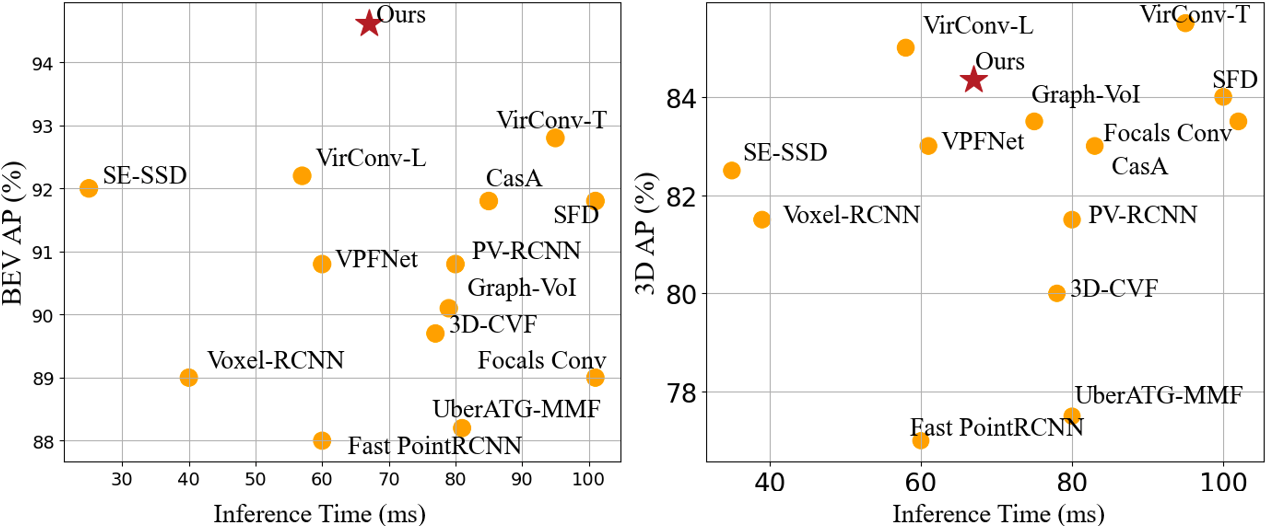}
\vspace{-6mm}
\caption{Comparison of inference time (ms) against  object detection accuracy (AP) on the KITTI dataset. MuStD, marked as red star, achieves superior accuracy  while maintaining competitive inference speed.}
\label{fig:compare}
\vspace{-3mm}
\end{figure}

\vspace{-4mm}

,
\section{Experiments}

We evaluated our MuStD network on the popular KITTI 3D object detection dataset~\cite{Geiger2012CVPR}, which is well-suited to autonomous driving research since it contains cars. The dataset comprises 7,481 training and 7,518 test samples, along with their corresponding LiDAR point clouds. Our experiments follow the standard KITTI protocol, using the specified sequences for training, validation, and testing. The model was trained for 40 epochs with a batch size of 2 and a learning rate of 1e-4, using a single RTX4090 GPU. Accuracy metrics were obtained by submitting the predictions on the test set to the official KITTI online evaluation server, ensuring standardized and transparent performance assessment. The  \href{https://www.cvlibs.net/datasets/kitti/eval_object_detail.php?&result=d162ec699d6992040e34314d19ab7f5c217075e0}{KITTI server results}
provide precision-recall curves, average precision (AP), and average orientation similarity (AOS) for evaluation. The server reports results across various benchmarks, including 2D and 3D detection, orientation estimation, and bird's-eye view (BEV) detection. Additionally, we conducted an ablation study on the validation set to analyze the contributions of various network components. We also evaluated our method for multi-class object detection. Our method was also evaluated for multi-class object detection. The proposed approach achieved outstanding results across multiple detection tasks, surpassing most existing methods in terms of accuracy and inference time. 

Table~\ref{tab:table6} summarizes the overall car detection results of our method for 2D, 3D, BEV, and orientation on the KITTI test set, as generated by the server. Figure~\ref{fig:compare} presents a comparison of inference time and detection accuracy for object detection methods on the KITTI dataset in both 3D detection and BEV. Our method demonstrates superior performance, achieving high accuracy in both tasks while maintaining competitive inference speed. Detailed results from the experiments are discussed in the following section.

\makeatletter
\def\thickhline{%
  \noalign{\ifnum0=`}\fi\hrule \@height \thickarrayrulewidth \futurelet
  \reserved@a\@xthickhline}
\def\@xthickhline{\ifx\reserved@a\thickhline
              \vskip\doublerulesep
              \vskip-\thickarrayrulewidth
             \fi
      \ifnum0=`{\fi}}
\makeatother

\begin{figure}
\includegraphics[height = 0.3 \columnwidth, width=1\columnwidth]{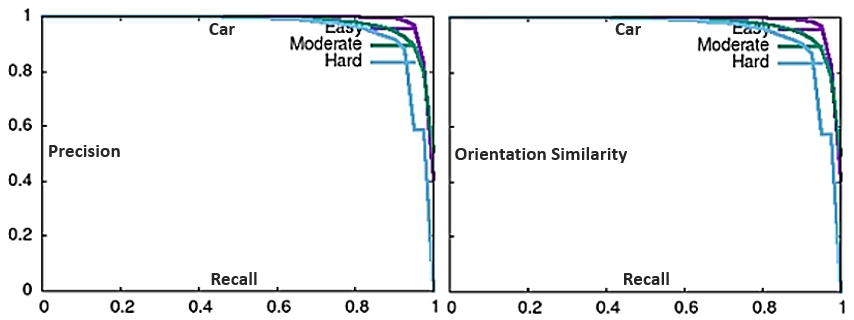}

\vspace{-3mm}
\caption{Car detection results of our method on KITTI test set for Easy, Moderate and Hard categories generated by the online KITTI server. Left: Recall-vs-precision curve for 2D car detection. Right: Recall-vs-orientation similarity curve for car orientation.}
\label{fig:main1}
\vspace{-4mm}
\end{figure}

\begin{figure}

\includegraphics[height = 0.3 \columnwidth, width=1\columnwidth]{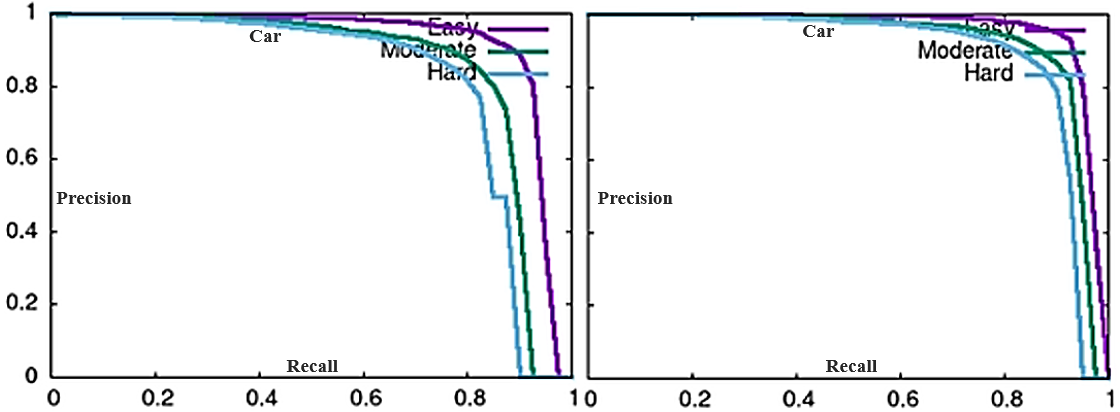}

\vspace{-3mm}
\caption{3D and BEV Car detection results of our method on the KITTI test set for Easy, Moderate and Hard categories generated by the online KITTI server. Left: Recall-vs-precision curve for 3D detection. Right: Recall-vs-precision curve for BEV detection.}
\label{fig:main2}
\vspace{-5mm}
\end{figure}

\vspace{-2mm}

\begin{table*}[t!]
\caption{Results on the KITTI object detection test set, generated by the online KITTI server. The average precision (AP) in \% is reported for car 3D and bird's-eye view (BEV) detection in all categories. Best results are bolded and 2nd best underlined.} 
\vspace{-2mm}
\centering
\setlength{\tabcolsep}{5pt}
\begin{tabular}{l |c |c |c c c c| c c c c |c}

\thickhline

 \multirow{2}{4em}{Approach} &  \multirow{2}{4em}{Reference} &  \multirow{2}{4em}{Modality}   &  \multicolumn{4}{c|}{Car 3D AP} &   \multicolumn{4}{c|}{Car BEV AP} &  \multirow{2}{4em}{Time (ms)}  \\   \cline{4-11}  
&  & & Easy & Moderate & Hard & mean AP &  Easy & Moderate & Hard & mean AP &  \\ \hline  \hline 

PV-RCNN~\cite{shi2020pv} & CVPR 2020 & LiDAR & 90.25 & 81.43 & 76.82 & 82.83 & 94.98 & 90.65 & 86.14 & 90.59 & 80* \\
Voxel-RCNN~\cite{deng2021voxel} & AAAI 2021 & LiDAR & 90.90 & 81.62 & 77.06 & 83.19 & 94.85 & 88.83 & 86.13 & 89.94 & 40 \\
CT3D~\cite{sheng2021improving} & ICCV 2021 & LiDAR & 87.83 & 81.77 & 77.16 & 82.25 & 92.36 & 88.83 & 84.07 & 88.42 & 70* \\
SE-SSD~\cite{zheng2021se} & CVPR 2021 & LiDAR & 91.49 & 82.54 & 77.15 & 83.73 & 95.68 & 91.84 & 86.72 & 91.41 & 30 \\
BtcDet~\cite{xu2022behind} & AAAI 2022 & LiDAR & 90.64 & 82.86 & 78.09 & 83.86 & 92.81 & 89.34 & 84.55 & 88.90 & 90 \\
CasA~\cite{casa2022} & TGRS 2022 & LiDAR & 91.58 & 83.06 & 80.08 & 84.91 & 95.19 & 91.54 & 86.82 & 91.18 & 86 \\
Graph-Po~\cite{yang2022graph} & ECCV 2022 & LiDAR & 91.79 & 83.18 & 77.98 & 84.32 & 95.79 & 92.12 & 87.11 & 91.01 & 60 \\\hline
berATG-MMF~\cite{liang2019multi} & CVPR 2019 & LiDAR+RGB & 88.40 & 77.43 & 70.22 & 78.02 & 93.67 & 88.21 & 81.99 & 87.29 & 80 \\
3D-CVF~\cite{yoo20203d} & ECCV 2020 & LiDAR+RGB & 89.20 & 80.05 & 73.11 & 80.79 & 93.52 & 89.56 & 82.45 & 88.51 & 75 \\
Focals Conv~\cite{focalsconv-chen} & CVPR 2022 & LiDAR+RGB & 90.55 & 82.28 & 77.59 & 83.47 & 92.67 & 89.00 & 86.33 & 89.33 & 100* \\
VPFNet~\cite{9826439} & TMM 2022 & LiDAR+RGB & 91.02 & 83.21 & 78.20 & 84.14 & 93.94 & 90.52 & 86.25 & 90.24 & 62 \\
Graph-VoI~\cite{yang2022graphrcnn} & ECCV 2022 & LiDAR+RGB & \textbf{91.89} & 83.27 & 77.78 & 84.31 & 95.69 & 90.10 & 86.85 & 90.88 & 76 \\
SFD~\cite{wu2022sparse} & CVPR 2022 & LiDAR+RGB & \underline{91.73} & \underline{84.76} & 77.92 & {84.80} & \textbf{95.64} & \underline{91.85} & 86.83 & \underline{91.44} & 98 \\
VirConv-L~\cite{VirConv} & CVPR 2023 & LiDAR+RGB & 91.41 & \textbf{85.05} & \underline{80.22} & \textbf{85.56} & \underline{95.53} & \textbf{91.95} & \underline{87.07} & \textbf{91.52} & 56 \\\hline \hline
Ours & - & LiDAR+RGB & {91.03} & 84.36 & \textbf{80.78} &\underline{85.39} & 94.62 & 91.13 & \textbf{88.28} & {91.34} & 67 \\
\thickhline
\end{tabular}
\label{tab:table2}
\vspace{-2mm}
\end{table*}


\begin{table*}[htbp]
\caption{Ablation study on  KITTI object detection validation set using different  fusion/component combinations of our  method. The average precision (AP) in \% is reported for 3D and 2D car detection. LiDAR-HC = LiDAR Height Compression stream.} 
\vspace{-2mm}
\centering
\setlength{\tabcolsep}{5pt}
\begin{tabular}{c|c| c | c | c | c c c c|c c c c}

\thickhline

 \multirow{2}{4em}{3D MM } &  \multirow{2}{5em}{LiDAR-HC}   &  \multirow{2}{6em}{LiDAR-Pillar } &  \multirow{2}{4em}{LiDAR} & \multirow{2}{2em}{RGB}& \multicolumn{4}{c|}{Car 3D AP} &   \multicolumn{4}{c}{Car 2D AP}    \\   \cline{6-13}  
&   &  & & & Easy & Moderate & Hard & mean AP &  Easy & Moderate & Hard & mean AP   \\ \hline 

 \checkmark & \checkmark & \checkmark & \checkmark & \checkmark & {95.21} & {93.56} & {90.09} & 92.95 &  98.42 & 97.75 & 94.07 & 96.08\\

 \checkmark & \checkmark &  & \checkmark & \checkmark & {92.77} & {90.19} & {87.30} & 90.75 &  95.11 & 95.81 & 92.27 & 94.39\\

 \checkmark &  & \checkmark & \checkmark & \checkmark & {91.33} & {89.14} & {86.15} & 88.87 &  94.67 & 93.53 & 90.70 & 92.97\\

  & \checkmark &  & \checkmark &  & {84.19} & {80.01} & {77.30} & 80.50 &  87.40 & 85.95 & 80.72 & 84.69\\

 \checkmark &  &  & \checkmark & \checkmark & {91.38} & {89.60} & {87.37} & 89.45 &  93.89 & 90.41 & 89.65 & 91.32\\

  & \checkmark & \checkmark & \checkmark &  & {88.48} & {86.07} & {83.93} & 86.83 &  91.06 & 89.34 & 86.73 & 89.04\\

\thickhline
\end{tabular}
\label{tab:table3}
\vspace{-5mm}
\end{table*}

\begin{table}[htbp]
    \centering
    \caption{Comparison with state-of-the-art on the KITTI validation set for 3D detection of multiple classes.}
\vspace{-2mm}
\setlength{\tabcolsep}{5pt}
\begin{tabular}{l |l |c c c c}

\hline 
 \multirow{2}{*} { \centering Class} &  \multirow{2}{*}{Method}  &  \multicolumn{4}{c}{3D Detection (AP in \%)}   \\  
&  & Easy & Moderate & Hard & mean \\ \hline  

 \multirow{3}{*} {Car} & Voxel-RCNN~\cite{deng2021voxel} & 89.39 &83.83 &87.73 & 86.32 \\
 & VirConv-T~\cite{VirConv} & \textbf{94.98} & 89.96 & 88.13 & 91.02 \\
 & Ours & {94.21} & \textbf{91.56} & \textbf{90.09} & \textbf{91.29} \\\hline

  \multirow{3}{*} {Pedestrian} & Voxel-RCNN~\cite{deng2021voxel} & 70.55 & 62.92 & 57.35 & 63.61 \\
 & VirConv-T~\cite{VirConv} & 73.32 & 66.93 & 60.38 & 66.88 \\
 & Ours & \textbf{75.45} & \textbf{68.11} & \textbf{63.40} & \textbf{68.32} \\\hline

  \multirow{3}{*} {Cyclist} & Voxel-RCNN~\cite{deng2021voxel} & 89.86 & 71.13 & 66.67 & 75.89 \\
 & VirConv-T~\cite{VirConv} & 90.04 & 73.90 & 69.06 & 77.00 \\
 & Ours & \textbf{91.89} & \textbf{76.56} & \textbf{71.91} & \textbf{80.12} \\\hline

\end{tabular}
\label{tab:table4}
\vspace{-7mm}
\end{table}

\subsection{2D car detection and orientation results on KITTI test set}
We evaluate the performance of MuStD on the KITTI test set for 2D car detection and orientation estimation. The KITTI server reports results for three difficulty levels: Easy, Moderate, and Hard, using AP for 2D detection and AOS for orientation similarity. 
Table~\ref{tab:table5} reports comparison of our method to the existing state-of-the-art. 
The proposed method outperforms all others in the Moderate and Hard categories for  2D detection as well car orientation. For the Easy category, our method achieves results very close to the top performer. Hence, the overall mean AP of our method is still the best for 2D car detection and car orientation.
These results underscore the effectiveness of our method, particularly in challenging (Hard and Moderate) scenarios with occluded or distant objects. 
Figure~\ref{fig:main1} visually illustrates the precision-recall and orientation similarity curves for the three difficulty levels. Our method is one of the fastest, with an inference time of only 50 milliseconds. Notice that our close competitors in accuracy, e.g., VirConv-T and SDF, are much slower.

\vspace{-0.7mm}
\subsection{3D and BEV detection results on KITTI Dataset}
\vspace{-0.7mm}
Table~\ref{tab:table2} compares our method for 3D and bird's eye view (BEV) detection on the KITTI test set (server-generated results) with the existing state-of-the-art. 
The proposed MuStD network achieves the best performance in the Hard category for 3D and BEV car detection. In the Easy and Moderate cases, our method achieves comparable results. 
For 3D detection, MuStD obtained 80.78\% AP in the Hard category and a mean AP of 85.3\% across all categories. For BEV, MuStD achieved 88.28\% AP in the Hard category and a mean AP of 91.34\%. 
These results highlight the effectiveness of our method in integrating LiDAR and RGB data and refining features through advanced mapping and indexing techniques. Precision-recall curves in Fig.~\ref{fig:main2} further demonstrate high precision across varying recall levels, reinforcing the efficacy of our multimodal fusion strategy.


\vspace{-1mm}

\subsection{Multi-class results on KITTI dataset}
\vspace{-1mm}
We also evaluate MuStD network on the KITTI validation set for multi-class 3D object detection. We test three classes that are most important for autonomous driving namely, `Car,' `Pedestrian,' and `Cyclist', across the three difficulty levels. Table~\ref{tab:table4} compares our results to Voxel-RCNN and VirConvT on the AP metric. Selection of these methods is based on thier performance and code availability. Our method consistently outperforms the competitors on all three categories achieving a mean AP of 91.29\% for Car, 68.32\% for Pedestrian, and 80.12\% for Cyclist. These results demonstrate our method's effectiveness in capturing geometric structures and spatial-texture features equally well for small and large objects in complex urban environments. Consistent performance across classes highlights the robustness and generalizability of our method in real-world scenarios.

\vspace{-2mm}
\subsection{Ablation study on KITTI validation set}
\vspace{-1mm}
\label{sec:abl}
To assess the impact of various components in the MuStD Network, we conducted an ablation study on the KITTI dataset, evaluating different configurations for 3D and 2D detection tasks. The study focused on key components: LiDAR-PillarNet, LiDAR Height Compression, and the 3D MM stream, along with LiDAR and RGB modalities. As summarized in Table~\ref{tab:table3}, the proposed MuStD achieved the best results with mean AP scores of 92.95\% for 3D detection and 96.08\% for 2D detection.
Removing the LiDAR-PillarNet stream led to about a 2\% drop in mean AP for both detection tasks, highlighting its role in capturing geometric details. Eliminating LiDAR Height Compression reduced AP to 88.87\% and 92.97\% for 3D and 2D detection, demonstrating its role in preserving spatial characteristics. Excluding the 3D MM Network resulted in a significant performance decline, with AP dropping to 80.50\% and 84.69\%, underscoring the necessity of integrating RGB and LiDAR data. LiDAR-only configurations yielded notably lower scores, confirming the critical role of multimodal fusion. These results emphasize the importance of combining all components for optimal detection performance.


\vspace{-2mm}

\section{Conclusion}
\vspace{-1mm}
We introduced the Multistream Detection Network (MuStD), a novel approach that enhances 3D object detection by integrating LiDAR point cloud data with RGB image information. Utilizing UV mapping and polar coordinate indexing, our method improves the extraction of geometric and spatial-texture features through a unified 2D representation. The network consists of three parallel streams—namely, the LiDAR-PillarNet Stream, the LiDAR Height Compression Stream, and the 3D MM Stream—which demonstrated superior performance on the popular KITTI benchmark. Experimental results highlight MuStD's high accuracy and robustness across various detection tasks, validating its effectiveness in addressing challenges related to detection loss, computational efficiency, and multimodal feature fusion. This work marks a significant advancement in 3D object detection for autonomous driving, providing a reliable solution for real-time navigation in complex urban environments.

\section*{Acknowledgment}
\vspace{-1mm}
Professor Ajmal Mian is the recipient of an Australian Research Council Future Fellowship Award (project number FT210100268) funded by the Australian Government. Dr.~Naveed Akhtar is the recipient of an Office of National Intelligence National Intelligence Postdoctoral Grant (project number NIPG-2021-001) funded by the Australian Government.

\vspace{-2mm}

\balance
\bibliographystyle{./IEEEtran} 
\bibliography{./IEEEabrv,./IEEEexample}

\begin{thebibliography}{10}
\providecommand{\url}[1]{#1}
\csname url@rmstyle\endcsname
\providecommand{\newblock}{\relax}
\providecommand{\bibinfo}[2]{#2}
\providecommand\BIBentrySTDinterwordspacing{\spaceskip=0pt\relax}
\providecommand\BIBentryALTinterwordstretchfactor{4}
\providecommand\BIBentryALTinterwordspacing{\spaceskip=\fontdimen2\font plus
\BIBentryALTinterwordstretchfactor\fontdimen3\font minus \fontdimen4\font\relax}
\providecommand\BIBforeignlanguage[2]{{%
\expandafter\ifx\csname l@#1\endcsname\relax
\typeout{** WARNING: IEEEtran.bst: No hyphenation pattern has been}%
\typeout{** loaded for the language `#1'. Using the pattern for}%
\typeout{** the default language instead.}%
\else
\language=\csname l@#1\endcsname
\fi
#2}}

\bibitem{Chen2017MV3D}
X.~Chen, H.~Ma, J.~Wan, B.~Li, and T.~Xia, ``Multi-view 3d object detection network for autonomous driving,'' in \emph{Proceedings of the IEEE Conference on Computer Vision and Pattern Recognition (CVPR)}, 2017, pp. 1907--1915.

\bibitem{arnold2019survey}
E.~Arnold, O.~Y. Al-Jarrah, M.~Dianati, S.~Fallah, D.~Oxtoby, and A.~Mouzakitis, ``A survey on 3d object detection methods for autonomous driving applications,'' \emph{IEEE Transactions on Intelligent Transportation Systems}, vol.~20, no.~10, pp. 3782--3795, 2019.

\bibitem{prakash2021multi}
A.~Prakash, K.~Chitta, and A.~Geiger, ``Multi-modal fusion transformer for end-to-end autonomous driving,'' in \emph{CVPR}, 2021, pp. 7077--7087.

\bibitem{wu2022sparse}
X.~Wu, L.~Peng, H.~Yang, L.~Xie, C.~Huang, C.~Deng, H.~Liu, and D.~Cai, ``Sparse fuse dense: Towards high quality 3d detection with depth completion,'' in \emph{CVPR}, 2022.

\bibitem{wu2022sfd}
B.~Wu, S.~He, Z.~Yan, W.~Zeng, and L.~Zhang, ``Sparse fuse dense: Towards high quality 3d detection with depth completion,'' in \emph{Proceedings of the IEEE/CVF Conference on Computer Vision and Pattern Recognition (CVPR)}, 2022, pp. 5412--5421.

\bibitem{wu2023virtual}
H.~Wu, C.~Wen, S.~Shi, X.~Li, and C.~Wang, ``Virtual sparse convolution for multimodal 3d object detection,'' in \emph{Proceedings of the IEEE/CVF Conference on Computer Vision and Pattern Recognition (CVPR)}, 2023, pp. 21\,653--21\,662.

\bibitem{9826439}
W.~Wang, J.~Shen, Z.~Wu, T.~He, J.~Zhang, Z.~Jiang, and G.~H. Lee, ``Vpfnet: Virtual point based feature fusion network for 3d object detection,'' in \emph{IEEE Transactions on Multimedia}, vol.~24, 2022, pp. 3487--3497.

\bibitem{yang2022graph}
Y.~Yang, X.~Sun, Z.~Zhang, K.~Jia, and W.~Zeng, ``Graph-voi: Graph neural network based voxel information aggregation for 3d object detection,'' in \emph{Proceedings of the European Conference on Computer Vision (ECCV)}, 2022, pp. 678--695.

\bibitem{zhu2023vpfnet}
H.~Zhu, J.~Deng, Y.~Zhang, J.~Ji, Q.~Mao, H.~Li, and Y.~Zhang, ``Vpfnet: Improving 3d object detection with virtual point based lidar and stereo data fusion,'' \emph{IEEE Transactions on Multimedia}, vol.~25, 2023.

\bibitem{bai2022transfusion}
X.~Bai, Z.~Hu, X.~Zhu, Q.~Huang, Y.~Chen, H.~Fu, and C.-L. Tai, ``Transfusion: Robust lidar-camera fusion for 3d object detection with transformers,'' in \emph{Proceedings of the IEEE/CVF Conference on Computer Vision and Pattern Recognition (CVPR)}, 2022, pp. 1090--1099.

\bibitem{guan2021m3detr}
T.~Guan, J.~Wang, S.~Lan, R.~Chandra, Z.~Wu, L.~Davis, and D.~Manocha, ``M3detr: Multi-representation, multi-scale, mutual-relation 3d object detection with transformers,'' \emph{arXiv preprint arXiv:2104.11896}, 2021.

\bibitem{meng2024efficient}
H.~Meng, C.~Li, G.~Chen, L.~Chen, and A.~Knoll, ``Efficient 3d object detection based on pseudo-lidar representation,'' \emph{IEEE Transactions on Intelligent Vehicles}, vol.~9, no.~1, pp. 1953--1964, 2024.

\bibitem{tian2023acfnet}
Y.~Tian, X.~Zhang, X.~Wang, J.~Xu, J.~Wang, R.~Ai, W.~Gu, and W.~Ding, ``Acf-net: Asymmetric cascade fusion for 3d detection with lidar point clouds and images,'' \emph{IEEE Transactions on Intelligent Vehicles}, vol.~9, no.~2, 2023.

\bibitem{wang2021multi}
D.~Wang, X.~Cui, X.~Chen, Z.~Zou, T.~Shi, S.~Salcudean, Z.~J. Wang, and R.~Ward, ``Multi-view 3d reconstruction with transformer,'' \emph{arXiv preprint arXiv:2103.12957}, 2021.

\bibitem{caesar2020nuscenes}
H.~Caesar, V.~Bankiti, A.~H. Lang, S.~Vora, V.~E. Liong, Q.~Xu, A.~Krishnan, Y.~Pan, G.~Baldan, and O.~Beijbom, ``nuscenes: A multimodal dataset for autonomous driving,'' in \emph{CVPR}, 2020, pp. 11\,621--11\,631.

\bibitem{8943956}
L.~Ma, Y.~Li, J.~Li, W.~Tan, Y.~Yu, and M.~A. Chapman, ``Multi-scale point-wise convolutional neural networks for 3d object segmentation from lidar point clouds in large-scale environments,'' \emph{IEEE Transactions on Intelligent Transportation Systems}, vol.~22, no.~2, pp. 821--836, 2021.

\bibitem{ICRAencode}
Y.~Zhang, L.~Wang, C.~Fu, Y.~Dai, and J.~M. Dolan, ``Encode: a deep point cloud odometry network,'' in \emph{2021 IEEE International Conference on Robotics and Automation (ICRA)}, 2021, pp. 14\,375--14\,381.

\bibitem{ICRAmatching}
A.~D. Pon, J.~Ku, C.~Li, and S.~L. Waslander, ``Object-centric stereo matching for 3d object detection,'' in \emph{2020 IEEE International Conference on Robotics and Automation (ICRA)}, 2020, pp. 8383--8389.

\bibitem{ICRAlidar}
J.~Fang, D.~Zhou, J.~Zhao, C.~Wu, C.~Tang, C.-Z. Xu, and L.~Zhang, ``Lidar-cs dataset: Lidar point cloud dataset with cross-sensors for 3d object detection,'' in \emph{2024 IEEE International Conference on Robotics and Automation (ICRA)}, 2024, pp. 14\,822--14\,829.

\bibitem{ICRAframe}
X.~Li, F.~Wang, N.~Wang, and C.~Ma, ``Frame fusion with vehicle motion prediction for 3d object detection,'' in \emph{2024 IEEE International Conference on Robotics and Automation (ICRA)}, 2024, pp. 4252--4258.

\bibitem{CVPR2021_Offboard3D}
C.~R. Qi, Y.~Zhou, M.~Najibi, P.~Sun, K.~Vo, B.~Deng, and D.~Anguelov, ``Offboard 3d object detection from point cloud sequences,'' in \emph{Proceedings of the IEEE/CVF Conference on Computer Vision and Pattern Recognition (CVPR)}, 2021, pp. 6134--6144.

\bibitem{Lang2019PointPillars}
A.~H. Lang, S.~Vora, H.~Caesar, L.~Zhou, J.~Yang, and O.~Beijbom, ``Pointpillars: Fast encoders for object detection from point clouds,'' in \emph{CVPR}, 19, pp. 12\,697--12\,705.

\bibitem{Yan2018SECOND}
Y.~Yan, Y.~Mao, and B.~Li, ``Second: Sparsely embedded convolutional detection,'' in \emph{Sensors}, vol.~18, no.~10, 2018, p. 3337.

\bibitem{shi2020pv}
S.~Shi, C.~Guo, L.~Jiang, Z.~Wang, J.~Shi, X.~Wang, and H.~Li, ``Pv-rcnn: Point-voxel feature set abstraction for 3d object detection,'' in \emph{Proceedings of the IEEE/CVF conference on computer vision and pattern recognition}, 2020, pp. 10\,529--10\,538.

\bibitem{deng2021voxel}
J.~Deng, S.~Shi, P.~Li, W.~Zhou, Y.~Zhang, and H.~Li, ``Voxel r-cnn: Towards high performance voxel-based 3d object detection,'' in \emph{Proceedings of the AAAI conference on artificial intelligence}, vol.~35, no.~2, 2021, pp. 1201--1209.

\bibitem{ji2022vision}
T.~Ji and L.~Xie, ``Vision-aided localization and navigation for autonomous vehicles,'' in \emph{2022 IEEE 17th International Conference on Control \& Automation (ICCA)}.\hskip 1em plus 0.5em minus 0.4em\relax IEEE, 2022, pp. 599--604.

\bibitem{Ku2018AVOD}
J.~Ku, M.~Mozifian, J.~Lee, A.~Harakeh, and S.~Waslander, ``Joint 3d proposal generation and object detection from view aggregation,'' in \emph{Proceedings of the IEEE/RSJ International Conference on Intelligent Robots and Systems (IROS)}, 2018, pp. 1--8.

\bibitem{yoo20203d}
J.~Yoo, J.~Kim, S.~Lee, M.~Roh, K.~M. Choi, and T.-K. Choi, ``3d-cvf: Generating joint camera and lidar features for robust 3d object detection,'' in \emph{Proceedings of the European Conference on Computer Vision (ECCV)}, 2020, pp. 282--300.

\bibitem{VirConv}
H.~Wang, J.~Li, K.~Zhang, and Y.-X. Wang, ``Virtual convolution for lidar-based 3d object detection,'' in \emph{Proceedings of the IEEE/CVF Conference on Computer Vision and Pattern Recognition (CVPR)}, 2023, pp. 13\,511--13\,520.

\bibitem{Sensors2024}
L.~Gao, H.~Xiang, X.~Xia, and J.~Ma, ``Multisensor fusion for vehicle-to-vehicle cooperative localization with object detection and point cloud matching,'' \emph{IEEE Sensors Journal}, vol.~24, no.~7, pp. 10\,865--10\,877, 2024.

\bibitem{sensors2022}
\BIBentryALTinterwordspacing
S.~Y. Alaba and J.~E. Ball, ``A survey on deep-learning-based lidar 3d object detection for autonomous driving,'' \emph{Sensors}, vol.~22, no.~24, 2022. [Online]. Available: \url{https://www.mdpi.com/1424-8220/22/24/9577}
\BIBentrySTDinterwordspacing

\bibitem{wang2023nrconv}
H.~Wang, B.~Li, X.~Song, H.~Li, and M.~Liu, ``Nrconv: Noise-resistant convolution for point cloud processing,'' in \emph{Proceedings of the IEEE/CVF Conference on Computer Vision and Pattern Recognition (CVPR)}, 2023, pp. 12\,011--12\,020.

\bibitem{ye2021twopass}
Y.~Ye, Y.~Wang, X.~Yang, S.~Wang, Z.~Huang, B.~Feng, and K.~Jia, ``2dpass: 2d priors assisted semantic segmentation of 3d scenes,'' in \emph{Proceedings of the IEEE/CVF Conference on Computer Vision and Pattern Recognition (CVPR)}, 2021, pp. 1609--1618.

\bibitem{zhou2023ted}
H.~Zhou, X.~Wang, L.~Chen, X.~Luo, H.~Zhang, and K.~Wu, ``Ted: Transformation-equivariant 3d detector,'' in \emph{Proceedings of the AAAI Conference on Artificial Intelligence (AAAI)}, 2023, pp. 2434--2442.

\bibitem{jhaldiyal2023semantic}
A.~Jhaldiyal and N.~Chaudhary, ``Semantic segmentation of 3d lidar data using deep learning: a review of projection-based methods,'' \emph{Applied Intelligence}, vol.~53, pp. 6844--6855, 2023.

\bibitem{li2023tinypillarnet}
Y.~Li, Y.~Zhang, and R.~Lai, ``Tinypillarnet: Tiny pillar-based network for 3d point cloud object detection at edge,'' \emph{IEEE Transactions on Circuits and Systems for Video Technology}, vol.~34, no.~3, 2023.

\bibitem{Choi_2019_ICCV}
H.~M. Choi, H.~Kang, and Y.~Hyun, ``Multi-view reprojection architecture for orientation estimation,'' in \emph{The IEEE International Conference on Computer Vision (ICCV) Workshops}, Oct 2019.

\bibitem{pang2020CLOCs}
S.~Pang, D.~Morris, and H.~Radha, ``Clocs: Camera-lidar object candidates fusion for 3d object detection,'' in \emph{2020 IEEE/RSJ International Conference on Intelligent Robots and Systems (IROS)}.\hskip 1em plus 0.5em minus 0.4em\relax IEEE, 2020.

\bibitem{ye2021spanet}
Y.~Ye, ``Spanet: Spatial and part-aware aggregation network for 3d object detection,'' in \emph{Pacific Rim International Conference on Artificial Intelligence}.\hskip 1em plus 0.5em minus 0.4em\relax Springer, 2021, pp. 308--320.

\bibitem{mao2021votr}
J.~Mao, Y.~Xue, M.~Niu, H.~Bai, J.~Feng, X.~Liang, H.~Xu, and C.~Xu, ``Voxel transformer for 3d object detection,'' in \emph{ICCV}, 2021.

\bibitem{mao2021pyramid}
J.~Mao, M.~Niu, H.~Bai, X.~Liang, H.~Xu, and C.~Xu, ``Pyramid r-cnn: Towards better performance and adaptability for 3d object detection,'' in \emph{ICCV}, 2021.

\bibitem{kim20223d}
Y.~Kim, K.~Park, M.~Kim, D.~Kum, and J.~W. Choi, ``3d dual-fusion: Dual-domain dual-query camera-lidar fusion for 3d object detection,'' \emph{arXiv preprint arXiv:2211.13529}, 2022.

\bibitem{casa2022}
H.~Wu, J.~Deng, C.~Wen, X.~Li, and C.~Wang, ``Casa: A cascade attention network for 3d object detection from lidar point clouds,'' \emph{IEEE Transactions on Geoscience and Remote Sensing}, 2022.

\bibitem{yang2022graphrcnn}
H.~Yang, Z.~Liu, X.~Wu, W.~Wang, W.~Qian, X.~He, and D.~Cai, ``Graph r-cnn: Towards accurate 3d object detection with semantic-decorated local graph,'' in \emph{ECCV}, 2022.

\bibitem{mahmoud2022dense}
A.~Mahmoud, J.~S. Hu, and S.~L. Waslander, ``Dense voxel fusion for 3d object detection,'' \emph{WACV}, 2023.

\bibitem{zhou2023octr}
C.~Zhou, Y.~Zhang, J.~Chen, and D.~Huang, ``Octr: Octree-based transformer for 3d object detection,'' in \emph{CVPR}, 2023.

\bibitem{focalsconv-chen}
Y.~Chen, Y.~Li, X.~Zhang, J.~Sun, and J.~Jia, ``Focal sparse convolutional networks for 3d object detection,'' in \emph{Proceedings of the IEEE Conference on Computer Vision and Pattern Recognition}, 2022.

\bibitem{TED}
H.~Wu, C.~Wen, W.~Li, R.~Yang, and C.~Wang, ``Transformation-equivariant 3d object detection for autonomous driving,'' in \emph{AAAI}, 2023.

\bibitem{lin2023mlf}
Z.~Lin, Y.~Shen, S.~Zhou, S.~Chen, and N.~Zheng, ``Mlf-det: Multi-level fusion for cross- modal 3d object detection,'' in \emph{International Conference on Artificial Neural Networks}.\hskip 1em plus 0.5em minus 0.4em\relax Springer, 2023, pp. 136--149.

\bibitem{yang2023pvt-ssd}
H.~Yang, W.~Wang, M.~Chen, B.~Lin, T.~He, H.~Chen, X.~He, and W.~Ouyang, ``Pvt-ssd: Single-stage 3d object detector with point-voxel transformer,'' in \emph{CVPR}, 2023.

\bibitem{Geiger2012CVPR}
A.~Geiger, P.~Lenz, and R.~Urtasun, ``Are we ready for autonomous driving? the kitti vision benchmark suite,'' in \emph{Conference on Computer Vision and Pattern Recognition (CVPR)}, 2012.

\bibitem{sheng2021improving}
H.~Sheng, S.~Cai, Y.~Liu, B.~Deng, J.~Huang, X.-S. Hua, and M.-J. Zhao, ``Improving 3d object detection with channel-wise transformer,'' in \emph{Proceedings of the IEEE/CVF International Conference on Computer Vision}, 2021, pp. 2743--2752.

\bibitem{zheng2021se}
W.~Zheng, W.~Tang, L.~Jiang, and C.-W. Fu, ``Se-ssd: Self-ensembling single-stage object detector from point cloud,'' in \emph{Proceedings of the IEEE/CVF conference on computer vision and pattern recognition}, 2021, pp. 14\,494--14\,503.

\bibitem{xu2022behind}
Q.~Xu, Y.~Zhong, and U.~Neumann, ``Behind the curtain: Learning occluded shapes for 3d object detection,'' in \emph{Proceedings of the AAAI Conference on Artificial Intelligence}, vol.~36, no.~3, 2022, pp. 2893--2901.

\bibitem{liang2019multi}
M.~Liang, B.~Yang, Y.~Chen, R.~Hu, and R.~Urtasun, ``Multi-task multi-sensor fusion for 3d object detection,'' in \emph{Proceedings of the IEEE/CVF conference on computer vision and pattern recognition}, 2019, pp. 7345--7353.

\end{thebibliography}
\end{document}